\documentclass[journal,twoside,web]{ieeecolor}
\usepackage{generic}
\usepackage{cite}
\usepackage{amsmath,amssymb,amsfonts}
\usepackage{graphicx}
\usepackage{booktabs}
\usepackage{hyperref}
\hypersetup{hidelinks=true}
\usepackage{textcomp}
\usepackage{xcolor}

\usepackage{orcidlink}

\def\BibTeX{{\rm B\kern-.05em{\sc i\kern-.025em b}\kern-.08em
    T\kern-.1667em\lower.7ex\hbox{E}\kern-.125emX}}

\makeatletter
\let\arxiv@includegraphics\includegraphics
\renewcommand{\includegraphics}[2][]{%
  \edef\arxiv@file{#2}\edef\arxiv@logo{\logoname.eps}%
  \ifx\arxiv@file\arxiv@logo\else\arxiv@includegraphics[#1]{#2}\fi}
\makeatother

\begin{document}

\title{Clinically Informed Colour Encoding Improves the Localisation
and Stability of Explanations in Deep Learning ECG Models: External
Validation in a UK Biobank Epilepsy Cohort}

\author{Lauren~Hutchinson$^{1}$\,\orcidlink{0009-0003-6073-2668}, Moses~Nkgari$^{1}$\,\orcidlink{0009-0003-7511-8564}, and~Alaa~Alahmadi$^{1,2}$\,\orcidlink{0000-0002-4846-116X}\\[4pt]
{\normalsize $^{1}$School of Computing, Newcastle University, Newcastle upon Tyne, UK\\
$^{2}$Department of Computer Science, The University of Manchester, Manchester, UK}
\thanks{This work has been submitted to the IEEE for possible publication. Copyright may be transferred without notice, after which this version may no longer be accessible.}
\thanks{This research was conducted using the UK Biobank Resource under Application Number 179597. Access to UK Biobank data was funded through The University of Manchester, with Newcastle University as a project partner.}
\thanks{\raggedright E-mails: L.Hutchinson4@newcastle.ac.uk;\\ M.Nkgari2@newcastle.ac.uk;\\ Alaa.Alahmadi@newcastle.ac.uk; Alaa.Alahmadi@manchester.ac.uk.}}

\maketitle

\begin{abstract}
\textit{Objective:} It is not known whether the input representation governs the quality of explanations of deep learning models for electrocardiogram analysis. Risk of dangerous ventricular repolarisation is judged from the QT-interval and the shape of the T-wave, features that are hard to assess visually and harder still to verify in a model's reasoning. We evaluate a pseudo-colour encoding filling the area beneath the QT interval, its colour sequence designed from clinical electrophysiology knowledge and mapped to a validated risk prediction model, the QT-nomogram. \textit{Methods:} Two identically configured residual networks were
trained on single-heartbeat images from a clinical trial, differing only in whether pseudo-colour encoding was applied, so that any difference is attributable to the chromatic channel alone. To externally validate the effect, both models were applied without fine-tuning or threshold tuning to a previously unseen epilepsy cohort from the UK-Biobank at two
visits, and eight explanation methods compared on faithfulness,
localisation and stability. \textit{Results:} The pseudo-colour model achieved areas under the
receiver operating characteristic curve of 0.903 and 0.914 at the first
and follow-up visits, against 0.719 and 0.720 for the monochrome model,
recovering three to five times as many at-risk heartbeats at comparable
specificity. Pseudo-colour encoding improved explanation localisation
and stability for almost every method, and the improvement reproduced across both visits.
\textit{Conclusion:} The input representation is a controllable and
previously overlooked determinant of both predictive performance and
explanation quality.
\textit{Significance:} Treating interpretability as a
representation-design problem offers a route towards explanations
clinicians can check against known electrophysiology.
\end{abstract}


\section{Introduction}

\IEEEPARstart{V}{entricular} depolarisation and repolarisation electrical activity appears on the surface electrocardiogram (ECG) as the interval from QRS onset to T-wave offset, known clinically as the QT interval. Prolongation of this interval reflects delayed repolarisation, which increases dispersion of repolarisation across the ventricular wall and thus the substrate for re-entrant fatal arrhythmia, of which Torsades de Pointes (TdP) is the most consequential form~\cite{yap2003drug,schwartz2016predicting}.

Duration alone, however, discriminates risk poorly~\cite{schwartz2016predicting}. Selective blockade of the rapid delayed rectifier potassium current prolongs repolarisation
and increases transmural dispersion, whereas agents additionally
inhibiting late sodium or calcium currents prolong the interval while
suppressing the arrhythmic substrate~\cite{Johannesen_2014}. The
discriminating information resides substantially in T-wave morphology
rather than QT-interval length~\cite{Johannesen_2014,bhuiyan2018history,sugrue2015electrocardiographic}

This matters for populations in whom repolarisation abnormality is
common and hard to assess. Sudden unexpected death accounts for a
substantial share of epilepsy-related mortality~\cite{Manolis_2019}, and
a proportion of patients exhibit cardiac electrical abnormality
including prolonged repolarisation~\cite{Manolis_2019,Sharma_2020}; a
pathogenic link between congenital long QT syndrome and seizure-like
presentation has been reported~\cite{Johnson_2009}. Epilepsy ECGs are
also noisier and more heterogeneous, making them a demanding test of any automated method.

Deep learning performs strongly on ECG tasks~\cite{Strodthoff_2021,
Jiang_2024} but its opacity limits adoption in safety-critical settings~\cite{Cina_2025}. Explanations, however, are not automatically trustworthy: attribution maps are unstable even across models of identical architecture, and there is no guarantee they land on structures a clinician would recognise as relevant~\cite{patlatzoglou2025cost}.

Morphology, however, resists the kind of verification that a duration measurement permits. An interval can be checked against a threshold; a shape cannot, and clinicians themselves disagree in assessing it. A deep learning model that predicts repolarisation risk accurately is therefore useful only if its reasoning can be inspected — and that inspection is only as good as the explanation it rests on. This paper asks whether the representation in which the electrocardiogram is presented to the network determines the quality of that explanation.

\subsection{Representation as an Interpretability Lever}

\noindent Encoding one-dimensional signals as images to exploit pretrained vision
architectures is established practice: Gramian angular fields, Markov
transition fields and recurrence plots have all been used this
way~\cite{Wang_2015,Hatami_2018}, and explanation methods have been
applied to such encodings as an inspection tool. Complementary work
argues the time domain is not always the most informative domain in
which to express an explanation, and constructs saliency in transformed
domains while holding the model fixed~\cite{TSSaliencyDomains}.
Sanity-check studies further show that input structure propagates into
attribution maps and that function smoothness affects saliency
quality~\cite{Kokhlikyan_2021}.

To our knowledge the input encoding has not been treated as a controlled
experimental variable with explanation quality as the dependent measure.
Existing comparisons between imaging transforms confound encoding with
geometry: a recurrence plot and a Gramian field differ in spatial
structure, information content and encoding simultaneously, so
differences in attribution cannot be attributed to any one. The present
design avoids this. Both representations are the same waveform rendering
at the same scale and position; only the chromatic channel differs.
Pseudo-colouring has been shown to enable lay observers to detect
prolonged repolarisation regardless of heart rate~\cite{Alahmadi_2020},
and, in the work immediately preceding this study, to improve few-shot
learning and produce clinically aligned explanations on the ECGRDVQ
dataset~\cite{Alahmadi_2025}. That work established the effect using a
single explanation method, evaluated qualitatively and without
validation beyond the source dataset. The present work evaluates eight
explanation methods against three quantitative criteria and tests
transfer to an unseen clinical UK Biobank cohort at two separate visits.

We therefore (i)~isolate the input representation as a determinant of
explanation quality using a matched-pair design; (ii)~evaluate eight
explanation methods against three quantitative criteria rather than by
inspection; (iii)~validate externally on an unseen epilepsy cohort; and
(iv)~propose a mechanistic account of why chromatic encoding aids
attribution, deriving a testable hypothesis about which component of
repolarisation the model attends to.

\section{Methods}

\subsection{Datasets}

\noindent For training, we used the ECGRDVQ (Electrocardiogram Ranolazine, Dofetilide, Verapamil, Quinidine) database, which is a publicly available clinical trial dataset comprising 5{,}235 12-lead ECG recordings from 22 healthy volunteers in a randomised crossover study of four QT-prolonging drugs vs placebo with differing ion-channel profiles~\cite{Johannesen_2014,Goldberger_2000}.
It was designed to demonstrate that interval duration alone
discriminates torsadogenic risk poorly, and so spans a controlled range of repolarisation morphologies of known pharmacological origin. Intervals were measured by expert cardiologists and risk labels assigned by the QT nomogram~\cite{Chan_2007}, giving 180 at-risk of TdP and 4{,}870
not-at-risk ECG recordings using lead II. Only one beat was retained per 10-second ECG recording
and all others discarded, so no two training images share a trace. No augmentation was applied.

For external validation, 12-lead resting ECG recordings (also using lead II) from participants with epilepsy were extracted from the UK Biobank~\cite{UKBiobank_2025} and used exclusively for
external evaluation. Epilepsy was ascertained using First
Occurrences fields (field identifiers 131048 and 131050), which record the earliest reported instance of a diagnosis mapped to ICD-10 codes G40 (epilepsy) and G41 (status epilepticus). Participants carrying either code and having an available electrocardiogram were included.

The 12-lead ECG is acquired at the imaging first visit
(instance~2) and the repeat imaging follow-up visit (instance~3); both were analysed, and are reported separately throughout.

Recordings were excluded where heart rate fell outside 40--100\,bpm or
where heart rate or the QT-interval ground truth values supplied by UK Biobank were missing.
The heart-rate range corresponds to that spanned by the training data (40--96\,bpm in ECGRDVQ), so evaluation is restricted to the regime in which the colour mapping was calibrated and over which the QT-nomogram assigns risk reliably. After exclusion, instance~2 comprised 873 participants (17 at risk of Torsades de Pointes (TdP) based on the QT-nomogram, 856 not at risk) and instance~3 comprised 168 participants (7 at risk, 161 not at risk).

All heartbeats within each 10-second trace were retained at test time for external validation, giving 8{,}217 single-heartbeat images at instance~2 (125 at risk, 8{,}092 not at risk) and 1{,}561 at instance~3 (49 at risk, 1{,}512 not at risk). Beat-level labelling is appropriate because repolarisation duration can vary between beats within a recording.

The resulting prevalence ($\approx\!1.5\%$ of beats at instance~2)
differs by more than an order of magnitude from that of the training data ($\approx\!27{:}1$), as do acquisition conditions: ECGRDVQ was recorded under controlled research conditions in healthy volunteers,
whereas the epilepsy cohort reflects routine clinical acquisition. The
low TdP event rate is clinically realistic for dangerous repolarisation
prolongation and is retained rather than corrected.

\subsection{Input Representations}

\begin{figure}[!t]
\centering
\includegraphics[width=0.85\columnwidth]{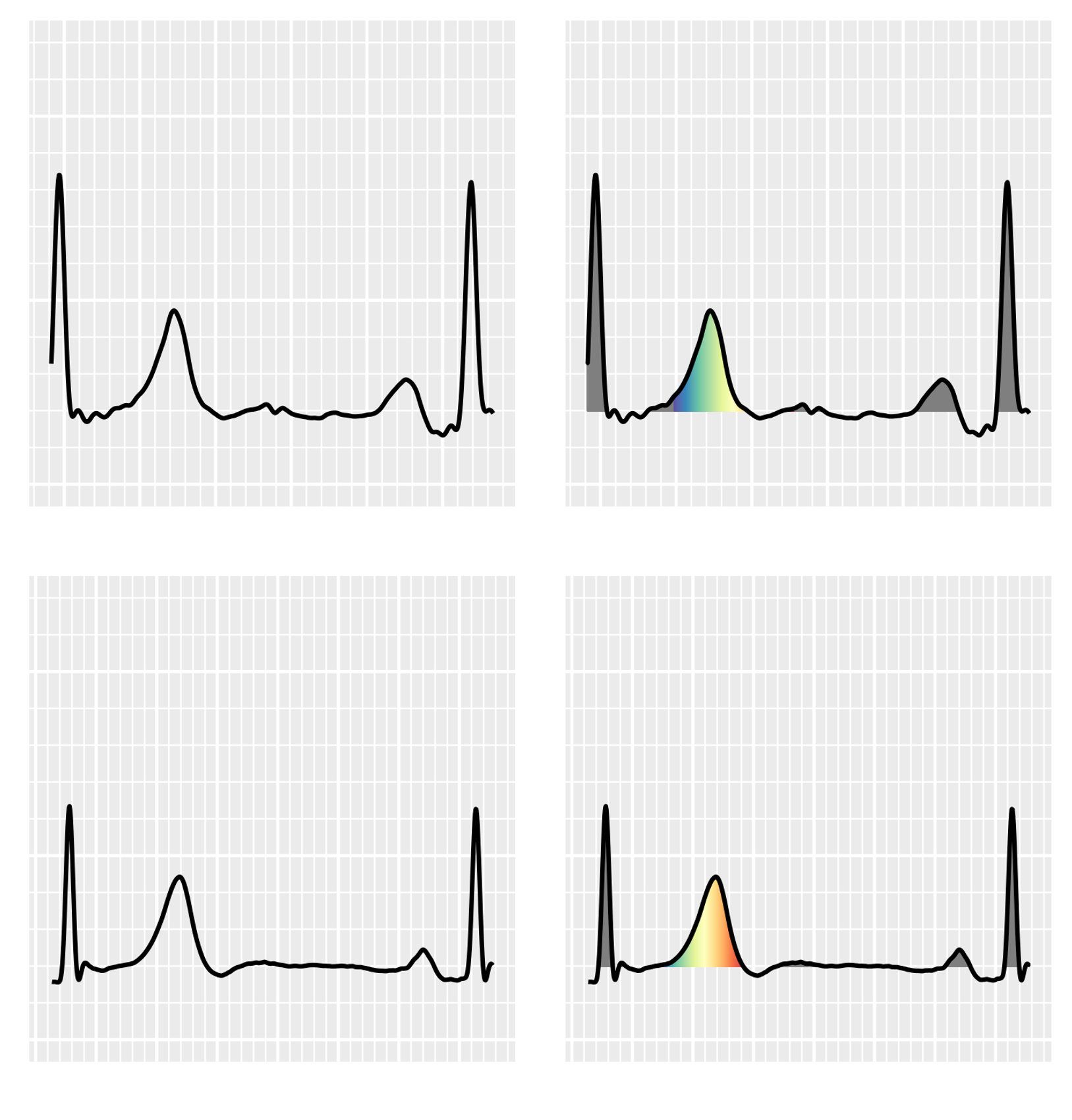}
\caption{Standard (left) and pseudo-colour (right) renderings of the
same beat, from a recording with prolonged ECG (bottom, HR = 45 and QT = 504ms), and normal ECG (top, HR =56 and QT=438ms). Geometry, scale and position are preserved exactly. Both have a sampling rate of 500 Hz, providing a resolution of 2 milliseconds per sample, making it difficult to perceive the risk visually without the colour (about a 66 ms subtle difference).}
\label{fig:representations}
\end{figure}

\noindent Each beat was rendered twice (Fig.~\ref{fig:representations}). The standard representation is a conventional monochrome ECG trace on a gridded background. The pseudo-colour representation fills the area beneath the QT interval, spanning the whole of ventricular repolarisation including the T-wave and, where present, the U-wave. The colour sequence is mapped according to the QT-nomogram~\cite{Chan_2007}, a clinically validated risk model, and is adjusted automatically to heart rate, so that the position of the waveform relative to the nomogram boundary is expressed chromatically~\cite{Alahmadi_2020}. The transform acts on the rendering and leaves the underlying signal unchanged; the two images are identical in geometry, scale and position, differing only in the chromatic channel.

The encoding is a deliberate injection of clinical prior knowledge into the input space. Rather than training an opaque model and explaining it afterwards, interpretability is embedded in the representation itself, so that human and machine reason over the same perceptually structured features~\cite{Alahmadi_2025}. Colour is a pre-attentive visual property, and mapping a temporally distributed quantity onto it renders visible a feature that is otherwise difficult to judge by eye~\cite{Alahmadi_2020}. Predictive performance under this representation should therefore be understood as the model exploiting encoded clinical knowledge rather than discovering repolarisation risk independently. Critically, however, the colour gradient spans the entire repolarisation area and does not specify where within that area diagnostic weight should fall; the distribution of attribution between early and late repolarisation remains free, and is the subject of the explanation analysis reported below.

\subsection{Why Encoding Should Affect Attribution}
\label{sec:mechanism}

\noindent A colour map adds no information: it is a deterministic function of the
signal already present. Any effect must therefore arise from how the
representation interacts with the network, not from what it encodes.

Two mechanisms are plausible. First, the classifier is initialised from
ImageNet weights, whose early filters encode colour-opponent and
textured natural-image statistics. A thin dark trace on a white ground
lies far outside that distribution and activates few pretrained filters,
whereas a chromatically encoded waveform lies closer to the manifold on
which they were learned. Second, and consequently, attribution is
computed on a different function. Gradient methods consume the local
derivative of the output with respect to the input; perturbation methods
consume the response surface under occlusion. Redistributing a
single-channel amplitude signal across three chromatic channels yields
denser, better-conditioned gradients where a monochrome trace is locally
flat. This is consistent with evidence that input structure propagates
into attribution maps and that function smoothness governs saliency
quality~\cite{Kokhlikyan_2021}.

This account predicts that a monochrome trace replicated across three
channels should behave like the standard arm, whereas an arbitrary
colour map with no clinical design should recover part but not all of
the benefit --- distinguishing it from the alternative account, that the
benefit arises specifically from clinical design.

\subsection{Model, Training and Explanation Methods}

\noindent Both models use a ResNet-18 backbone from ImageNet
weights~\cite{PyTorch_2026}, with \texttt{conv1}--\texttt{layer2} frozen
to retain transferable visual features~\cite{Yang_2023} and
\texttt{layer3}--\texttt{layer4} fine-tuned. The head was replaced with
a binary classifier preceded by dropout ($p\,{=}\,0.3$). Training used
Adam ($\eta\,{=}\,1\!\times\!10^{-4}$), batch size 32, step decay
(step 5, $\gamma\,{=}\,0.5$) and cross-entropy loss for 30 epochs on the
complete ECGRDVQ dataset, with no internal validation split and no early
stopping. Images were resized to $224\!\times\!224$ and normalised with
ImageNet statistics. The two runs were identical in architecture,
hyperparameters, data and random seed, differing only in representation;
determinism was enforced throughout by seed fixing, deterministic
kernels and disabled cuDNN benchmarking. Trained models were applied to
the epilepsy cohort without fine-tuning or domain adaptation for external validation.

Eight explanation methods were applied to both models using
Captum~\cite{Captum_2026}: Saliency, Guided Backpropagation, Integrated
Gradients (blurred-waveform baseline), GradientShap, DeepLIFT with
NoiseTunnel, layer-wise relevance propagation (LRP) and Grad-CAM
(gradient family) and Occlusion (perturbation).
Attention methods were excluded as they require architectural change
rather than post-hoc explanation of the same model, and attention
weights are not reliably faithful~\cite{Niu_2021}. Explanations were computed for every at-risk beat at both visits (125 beats from 17 participants at instance~2; 49 from seven participants at
instance~3), with both representations evaluated on the same beats to
permit paired comparison. LIME was not included, its stochastic sampling
and superpixel segmentation placing it on a different scale from the
methods above and rendering it sensitive to the chromatic difference
under test.

\subsection{Evaluation Protocol}

\noindent Discrimination is reported by the area under the receiver operating
characteristic curve (AUC-ROC), which is independent of any decision
threshold. Threshold-dependent metrics are reported at the default
operating point of 0.5; no threshold tuning was performed, so no
information from the evaluation cohorts entered the choice of operating
point. Balanced accuracy, the mean of sensitivity and specificity, is
reported in preference to raw accuracy, which at the prevalence observed
here is dominated by the majority class. Positive predictive value is
similarly governed principally by the base rate rather than by model
behaviour and is not reported.

Beats are clustered within participants, so the effective sample size
approaches the participant count rather than the beat count. Confidence
intervals were therefore obtained by bootstrap resampling of
\emph{participants}, with both models re-scored on each resample so that
the interval on the between-representation difference is itself paired.

\subsection{Explanation Quality Metrics}
\label{sec:metrics}

\noindent \emph{Faithfulness} measures whether an explanation reflects the model’s actual decision\-making. It was assessed by perturbing ECG regions identified as important and measuring the resulting change in predicted QT risk score. Larger decreases in the score indicate that the perturbed region was genuinely influential, meaning the explanation successfully captured clinically aligned waveform features the model relies on~\cite{Yeh_2019}.

\emph{Localisation} evaluates how well explanations focus on the masked QT region. The mask provides a simple baseline \cite{Montavon_2018} but only approximates QT morphology, so localisation results should be interpreted as alignment with general waveform structure rather than precise QT‑specific or clinically verified localisation.

\emph{Stability} measures how consistent an explanation is under small input perturbations. Gradient-based stability was computed using cosine similarity between repeated attributions \cite{Nguyen_2026}. More stable explanations are more reliable for clinical interpretation because they reflect genuine model behaviour rather than noise \cite{Tomsett_2020}.

Confidence intervals were obtained by non-parametric cluster bootstrap:
participants were resampled with replacement, all beats from each
selected participant retained, and the median recomputed, with intervals
taken as the 2.5th and 97.5th percentiles of 5{,}000 resamples.
Resampling at the participant rather than beat level accounts for the
correlation between beats from the same individual, which materially
widens the intervals for faithfulness.

\section{Results}

\subsection{Discrimination Under Domain Shift}

\noindent Applied without fine-tuning or threshold tuning to the epilepsy cohort, the pseudo-colour model achieved an AUC-ROC of 0.903 at instance~2 and 0.914 at instance~3, against 0.719 and 0.720 for the monochrome model (Table~\ref{tab:external}). The advantage of chromatic encoding was therefore approximately 0.19 in AUC-ROC and was reproduced almost exactly across the two visits.

The two representations differed asymmetrically in the two error types. Specificity was closely comparable between them at both visits (87.9\% against 90.6\% at instance~2; 89.6\% against 90.8\% at
instance~3), whereas sensitivity differed by roughly fifty percentage points (68.0\% against 18.4\%; 67.3\% against 14.3\%). At effectively matched false-positive rates, the pseudo-colour representation therefore recovered between three and five times as many at-risk beats: 85 of 125 against 23 at instance~2, and 33 of 49 against 7 at instance~3. The difference is not an artefact of the operating point, since the monochrome receiver operating characteristic curve rises shallowly across its full range rather than crossing the pseudo-colour curve.

Raw accuracy was equivalent between representations at both visits (88--89\%), which at this prevalence reflects the majority class rather than model behaviour: the monochrome model attains it while failing to identify 102 of 125 at-risk beats. Balanced accuracy, which weights the two classes equally, separates the representations clearly (78.0\% against 54.5\% at instance~2; 78.5\% against 52.5\% at instance~3) and places the monochrome model only marginally above chance.

Performance was closely comparable across the two instances for both representations, differing by 0.011 and 0.001 in AUC-ROC respectively. Because instance~3 comprises largely the same participants reassessed some years later, this consistency is best read as stability of model output within individuals over time rather than as independent replication. The instance~3 estimates rest on 49 at-risk beats from seven participants and are correspondingly imprecise.

At a beat-level prevalence of 1.5\%, the default operating point yields 983 false positives against 85 true positives for the pseudo-colour model at instance~2. This ratio follows from the base rate rather than from model behaviour, and is the reason threshold-free discrimination is
reported as the primary measure. For a screening application in which
flagged recordings would receive expert review rather than intervention,
a high false-positive rate at high sensitivity may be an acceptable
trade; the appropriate operating point is a deployment decision that
this evaluation does not attempt to fix.

\begin{table}[!t]
\caption{External evaluation on the UK Biobank epilepsy cohort at the default
operating point, with no threshold tuning. Instance~2 is the first imaging
visit and instance~3 the repeat imaging follow-up visit; the latter comprises
largely the same participants and is reported as a repeated measurement.
Positive predictive value is governed by the base rate at this
prevalence and is not reported.}
\label{tab:external}
\centering
\begin{tabular}{lcccc}
\toprule
& \multicolumn{2}{c}{Instance 2} & \multicolumn{2}{c}{Instance 3} \\
\cmidrule(lr){2-3}\cmidrule(lr){4-5}
Total participants & \multicolumn{2}{c}{873} & \multicolumn{2}{c}{168} \\
\quad at TdP risk & \multicolumn{2}{c}{17} & \multicolumn{2}{c}{7} \\
\midrule
Heartbeat images & \multicolumn{2}{c}{8{,}217} & \multicolumn{2}{c}{1{,}561} \\
\quad at TdP risk & \multicolumn{2}{c}{125} & \multicolumn{2}{c}{49} \\
\midrule
& Colour & Mono. & Colour & Mono. \\
\midrule
AUC-ROC & \textbf{0.903} & 0.719 & \textbf{0.914} & 0.720 \\
Sensitivity & \textbf{68.0\%} & 18.4\% & \textbf{67.3\%} & 14.3\% \\
Specificity & 87.9\% & 90.6\% & 89.6\% & 90.8\% \\
Balanced accuracy & \textbf{78.0\%} & 54.5\% & \textbf{78.5\%} & 52.5\% \\
\bottomrule
\end{tabular}
\end{table}

At a beat-level prevalence of 1.5\%, the default operating point yields
983 false positives against 85 true positives at instance~2. This ratio
follows from the base rate rather than from model behaviour, and is the
reason threshold-free discrimination is reported as the primary measure.
For a screening application in which flagged recordings would receive
expert review rather than intervention, a high false-positive rate at
high sensitivity may be an acceptable trade; the appropriate operating
point is a deployment decision that this evaluation does not attempt to
fix.

\subsection{Explanation Quality}

\noindent Explanation metrics were computed for every at-risk beat at both visits
(125 beats from 17 participants at instance~2; 49 beats from seven
participants at instance~3), with both representations evaluated on the
same beats so that all comparisons are paired
(Table~\ref{tab:xai}).

\emph{Localisation} was higher under pseudo-colour for seven of the
eight methods at instance~2 and for all eight at instance~3, with the
gain reaching $+0.115$ for Saliency and $+0.109$ for Integrated
Gradients. LRP was the sole exception, localising better under the
monochrome representation at instance~2 ($-0.089$) but not at
instance~3.

\emph{Stability} was higher under pseudo-colour for every method at both
visits. Gains were largest for the gradient family --- Guided
Backpropagation $+0.185$, Saliency $+0.179$, DeepLIFT $+0.166$ at
instance~2 --- and negligible for Occlusion and Grad-CAM, which are
near-deterministic under both representations by construction. LRP was
unstable under both (median $\le 0.010$ throughout), indicating
inconsistency arising from the method rather than from the input.

\emph{Faithfulness} was higher under pseudo-colour for every method at
both visits, and the differences were large: median $0.62$ against
$-0.12$ for Guided Backpropagation and $0.82$ against $-0.63$ for
Occlusion at instance~2. All differences survived correction
($p < 0.01$).

\subsection{Replication Across Visits}

\noindent The between-representation differences reproduced closely at the two
visits despite involving different participants, different beats and
recording sessions separated by several years. Localisation differences
agreed to within $0.02$ for seven of eight methods and reproduced to
three decimal places for Guided Backpropagation ($+0.092$ at both
visits) and GradientShap ($+0.037$ at both); stability differences
agreed to within $0.01$ for every method. Absolute values were similarly
consistent: under pseudo-colour, localisation differed between visits by
at most $0.017$ and stability by at most $0.006$ across all eight
methods.

Two method-level failures likewise reproduced. Grad-CAM achieved a
median faithfulness of $0.003$ at both visits, indicating that its
attributions carry essentially no information about the model's
dependence on the input at this resolution; computing attribution from a
coarse final feature map upsampled to input resolution is poorly suited
to a thin rendered waveform. LRP achieved a median stability of $0.006$
and $0.010$, so its localisation values should not be interpreted.

\begin{table*}[!t]
\caption{Explanation quality by method, representation and visit, for
at-risk beats. Values are medians; all colour--monochrome differences
are significant at $p<0.05$ after Benjamini--Hochberg correction across
methods except those marked $\dagger$. Both representations were
evaluated on the same beats, so comparisons are paired.}
\label{tab:xai}
\centering
\small
\begin{tabular}{lcccc|cccc|cccc}
\toprule
& \multicolumn{4}{c|}{Localisation} & \multicolumn{4}{c|}{Stability}
& \multicolumn{4}{c}{Faithfulness} \\
& \multicolumn{2}{c}{Instance 2} & \multicolumn{2}{c|}{Instance 3}
& \multicolumn{2}{c}{Instance 2} & \multicolumn{2}{c|}{Instance 3}
& \multicolumn{2}{c}{Instance 2} & \multicolumn{2}{c}{Instance 3} \\
Method & Col. & Mono & Col. & Mono & Col. & Mono & Col. & Mono
       & Col. & Mono & Col. & Mono \\
\midrule
Guided Backprop & 0.897 & 0.805 & 0.911 & 0.819 & 0.985 & 0.800 & 0.986 & 0.791 & 0.617 & $-$0.116 & 0.685 & $-$0.164 \\
Integrated Grad. & 0.801 & 0.692 & 0.800 & 0.699 & 0.739 & 0.601 & 0.742 & 0.602 & 0.628 & 0.014 & 0.616 & 0.027 \\
DeepLIFT & 0.700 & 0.634 & 0.706 & 0.635 & 0.741 & 0.575 & 0.746 & 0.586 & 0.591 & $-$0.146 & 0.639 & $-$0.154 \\
Saliency & 0.809 & 0.695 & 0.798 & 0.701 & 0.704 & 0.525 & 0.710 & 0.526 & 0.626 & 0.023 & 0.635 & 0.041 \\
GradientShap & 0.905 & 0.867 & 0.918 & 0.881 & 0.921 & 0.813 & 0.923 & 0.818 & 0.620 & $-$0.048 & 0.631 & $-$0.028 \\
Occlusion & 0.804 & 0.746 & 0.830 & 0.746 & 0.999 & 0.993 & 0.999 & 0.992 & 0.818 & $-$0.630 & 0.875 & $-$0.594 \\
Grad-CAM & 0.693 & 0.673 & 0.710$^\dagger$ & 0.683 & 0.999 & 0.996 & 1.000 & 0.996 & 0.003 & 0.000 & 0.003 & 0.000 \\
LRP & 0.736 & 0.825 & 0.792$^\dagger$ & 0.778 & 0.006$^\dagger$ & 0.008 & 0.010$^\dagger$ & 0.008 & 0.517 & $-$0.190 & 0.504 & $-$0.165 \\
\bottomrule
\end{tabular}
\end{table*}

\subsection{Faithfulness Tracks Prediction Outcome}

\noindent The faithfulness differences require qualification. Within the
pseudo-colour arm, faithfulness separated sharply according to whether
the model classified the beat correctly: at instance~2 the median for
correctly flagged beats was $0.698$ against $-0.092$ for missed beats
under Guided Backpropagation, with comparable separation for every
gradient method ($p < 10^{-17}$). The relationship was graded rather
than binary, faithfulness correlating with the predicted at-risk
probability at Spearman $\rho = 0.91$--$0.96$ across methods.

Because the monochrome model has low sensitivity and flagged far fewer at-risk beats than the pseudo-colour model (23 of 125 against 85 at instance~2; seven of 49 against 33 at instance~3), much of the apparent faithfulness advantage
follows from this difference in detection rather than from a property of
the explanations themselves. Restricting the comparison to the 20 beats
that \emph{both} models classified correctly, the localisation and
stability advantages persisted for six of seven methods
($p < 0.05$), whereas the faithfulness advantage did not, reversing in
direction for Integrated Gradients, Saliency and GradientShap. We
therefore regard localisation and stability as properties of the
representation, and the faithfulness difference as substantially a
consequence of the pseudo-colour model detecting cases that the
monochrome model misses.

\subsection{Qualitative Explanations}

\begin{figure*}[!t]
\centering
\includegraphics[width=\textwidth]{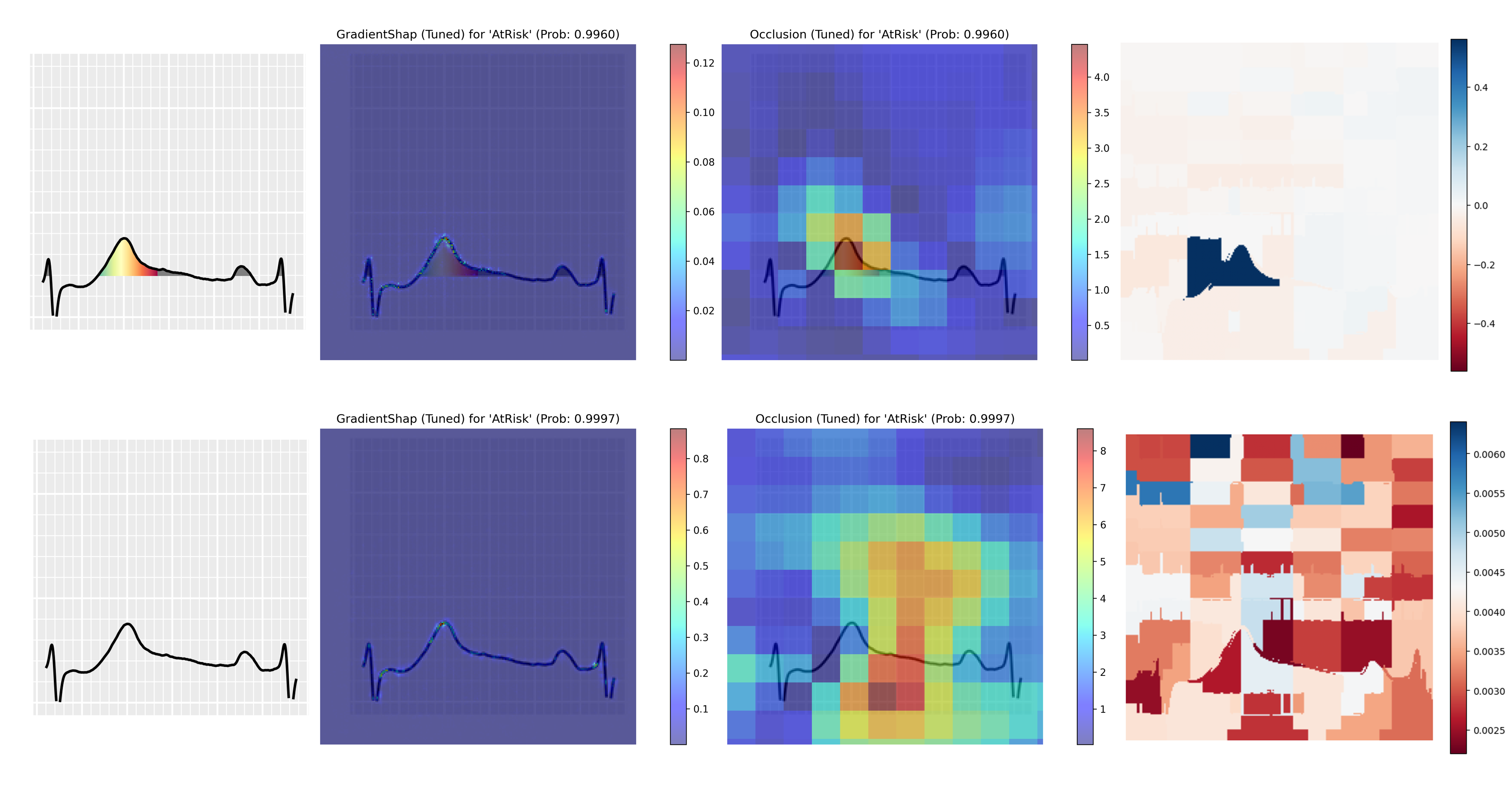}
\caption{Explanations for a correctly classified at-risk heartbeat under the pseudo-colour (top) and monochrome (bottom) representations, shown for GradientShap, Occlusion and LIME. The leftmost panel of each row shows the input: under pseudo-colour, the fill beneath the QT interval extends into the warm end of the sequence, indicating a repolarisation duration approaching the nomogram threshold. GradientShap attributes to the repolarisation segment under both representations but with low contrast against the surrounding waveform. Occlusion shows the clearest difference: attribution forms a compact region over the T wave under pseudo-colour, whereas under monochrome it spreads across a wider area extending into segments without diagnostic content.}
\label{fig:qualitative}
\end{figure*}

\noindent Representative explanations for a correctly classified at-risk beat are
shown in Fig.~\ref{fig:qualitative}. GradientShap attributes to the
repolarisation region under both representations but with low visual
contrast. Occlusion disperses attribution across regions without
diagnostic content under the standard representation and concentrates on
the repolarisation segment under pseudo-colour. Being drawn from
individual beats, these observations are illustrative rather than
evidential.

\begin{figure*}[!t]
\centering
\includegraphics[width=\textwidth]{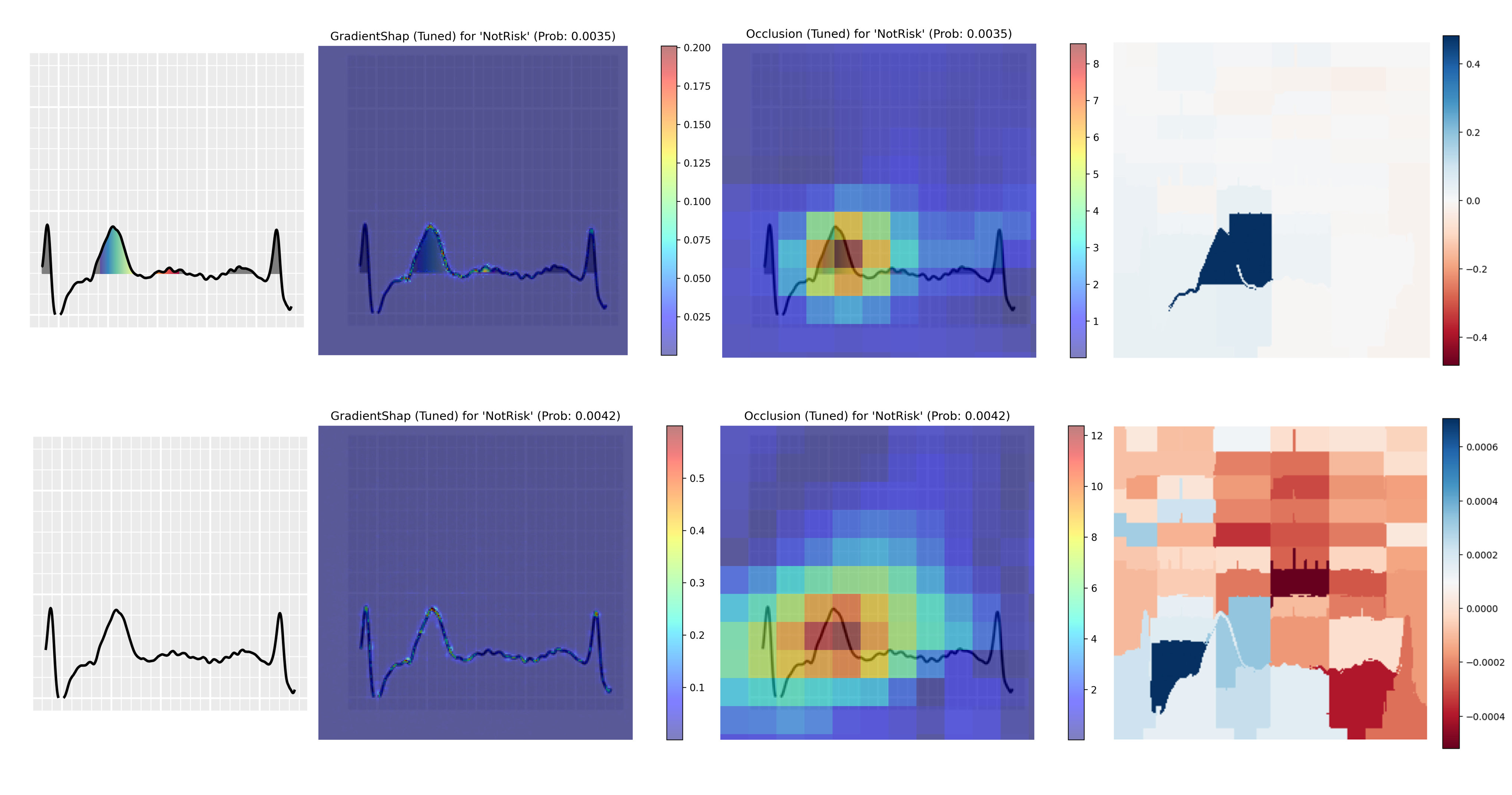}
\caption{Explanations for a correctly classified not-at-risk heartbeat, arranged as in Fig.~\ref{fig:qualitative}. The pseudo-colour input (top left) shows the fill terminating within the cool end of the sequence, consistent with a repolarisation duration well below the nomogram threshold. The pattern seen for the at-risk case is reproduced here in the not-at-risk case: GradientShap is diffuse under both representations; Occlusion concentrates over the T wave under pseudo-colour and disperses under monochrome; and LIME resolves a single region under pseudo-colour against fragmented attribution under monochrome. That the same contrast holds for a beat the model assigns to the low-risk class indicates the effect is a property of the representation rather than of the predicted class.}
\label{fig:qualitative_norisk}
\end{figure*}

\section{Discussion}

\noindent The representation in which a physiological signal is presented to a
network is a controllable determinant both of predictive performance and
of the quality of the explanations subsequently extracted from it.
Chromatic encoding of the repolarisation interval raised discrimination
by approximately 0.19 in AUC-ROC and improved the localisation and
stability of attribution for almost every explanation method examined.
Because the two representations are identical in geometry, scale and
position, these differences are attributable to the chromatic channel
rather than to confounded changes in spatial structure --- a control
that comparisons between imaging transforms cannot achieve.

The most substantial evidence is the replication. Effect sizes for the
between-representation difference reproduced across two visits involving
different participants, different beats and recording sessions separated
by years, agreeing to within 0.02 for localisation and 0.01 for
stability, and in two cases to three decimal places. Explanation quality
under this representation is therefore a stable property of the pairing
of method with representation rather than an artefact of a particular
sample.

The results also identify two method-level failures that reproduced at
both visits. Grad-CAM produced attributions with essentially no
faithfulness in this setting, and LRP produced attributions with
essentially no stability. Neither is a limitation of the encoding; both
indicate that method selection for rendered physiological signals cannot
be assumed from performance on natural images. Guided Backpropagation
and GradientShap performed best overall, combining high localisation
with high stability under pseudo-colour.

\subsection{Physiological Interpretation and Hypothesis}
\label{sec:physiology}

\noindent Risk labels in both cohorts derive from the QT nomogram, a criterion
defined on interval duration against heart rate; no component of the
training signal encodes T-wave shape. Attribution under pseudo-colour
nonetheless concentrated on the repolarisation segment rather than
distributing across the full depolarisation--repolarisation complex.

This is of interest given the provenance of the training data. ECGRDVQ
was constructed to show that duration alone discriminates torsadogenic
risk poorly: selective potassium-current blockade prolongs late
repolarisation and increases transmural dispersion, whereas multichannel
blockade prolongs the interval while suppressing the arrhythmic
substrate~\cite{Johannesen_2014}. The discriminating information resides
in terminal T-wave morphology, conventionally quantified from T-wave
peak to T-wave offset.

That attribution should favour this region, under a label not
referencing it, suggests the network may have acquired a morphological
correlate of repolarisation abnormality rather than an implicit interval
measurement.

We state this as a hypothesis. The explanation sample is small; the mask
approximates waveform structure rather than delineated boundaries; and
the colour map was designed to emphasise T-wave morphology, so
attribution there may partly reflect the encoding. Three analyses would
discriminate these accounts: recomputing localisation against masks from
automated delineation of QRS onset, T peak and T offset, permitting
early and late repolarisation to be quantified separately; substituting
an arbitrary colour map, which should degrade T-wave attribution if the
effect is knowledge-driven; and stratifying attribution by administered
agent within ECGRDVQ, testing whether selective and multichannel
blockers produce distinguishable late-repolarisation signatures.

In the epilepsy UK Biobank cohort, repolarisation abnormality may reflect autonomic modulation, chronic antiseizure medication exposure, or shared
ion-channel pathology between neuronal and cardiac tissue --- the last
consistent with seizure-like presentation among genotyped long QT
patients~\cite{Johnson_2009}. Attribution-based analysis of a cohort
with concurrent electroencephalography and genotyping would offer a
route to separating these, and is the natural extension towards
brain--heart interaction.

\subsection{Interpretation, Limitations and Scope}

\noindent The same encoding improves detection of prolonged repolarisation by
observers without cardiology training~\cite{Alahmadi_2020} and, as shown
here, the stability and localisation of machine attribution. A
representation built from clinical knowledge of where diagnostic
information resides thus aids both human and machine interpretation ---
which is what makes explanations in image space checkable, the
attribution being expressed over the same visual features a clinician
already uses. This motivates rendering rather than modelling the raw
signal: the colour encoding has no direct analogue in signal space,
clinical ECG interpretation is itself visual, and image space permits
transfer from large-scale pretraining, material given 22 training
individuals. Encodings such as recurrence plots, though also image-based,
would forfeit that clinical readability.

Limitations bear on interpretation. The pseudo-colour encoding embeds
the nomogram criterion from which the risk labels derive, so
discrimination under that representation demonstrates effective use of
injected clinical knowledge rather than independent discovery of
repolarisation morphology; the substantive result here concerns where
attribution falls within the encoded region. Training used 22
individuals (5{,}235 ECGs), so the learned representation may not span clinical
morphological variation. At the observed prevalence the default
operating point yields many false positives per true positive, and no
operating point is proposed for deployment. The explanation analysis
rests on at-risk beats from 17 and 7 participants respectively, so
effective sample size is closer to the participant than the beat count;
confidence intervals were computed accordingly. The localisation mask
approximates waveform structure rather than delineated repolarisation
boundaries. Finally, a colour-dependent representation raises questions
of accessibility for observers with colour-vision deficiency and of
fidelity under monochrome reproduction, neither assessed here.

\section{Conclusion}

\noindent Explanation quality in deep learning models of physiological signals
depends not only on the explanation method but on the representation in
which the signal is presented. Using a matched-pair design isolating the
chromatic channel, we find that clinically informed colour encoding
improves the stability of gradient-based attribution, modestly improves
localisation, reduces catastrophic explanation failure, and preserves
discrimination and probability calibration under domain shift better
than conventional monochrome rendering. Treating interpretability as a problem of representation design offers a route towards explanations clinicians can verify against established electrophysiology and can have wider applications beyond QT-prolongation.

\section*{References}

\bibliographystyle{IEEEtran}
\bibliography{references}

\end{document}